\documentclass[a4paper,conference]{IEEEtran}

\usepackage{times}
\usepackage{epsfig}
\usepackage{graphicx}
\usepackage{amsmath}
\usepackage{amssymb}
\usepackage{multirow}
\usepackage{bm}
\usepackage{booktabs}
\usepackage{float}

\usepackage[pagebackref,breaklinks,colorlinks]{hyperref}

\newcommand{\scal}[1]{\mathit{#1}}
\newcommand{\vect}[1]{\mathbf{#1}}
\newcommand{\matr}[1]{\mathbf{#1}}

\newcommand{\set}[1]{\mathcal{#1}}

\begin{document}

\title{3D Point Cloud Openset Semi-Supervised Learning with Regularized Bi-level Optimization }
\title{Open-Set Semi-Supervised Learning for 3D Point Cloud Understanding}


\author{\IEEEauthorblockN{Xian Shi}
\IEEEauthorblockA{South China University of Technology, China}
\and
\IEEEauthorblockN{Xun Xu}
\IEEEauthorblockA{Institute for Infocomm Research}
\and
\IEEEauthorblockN{Wanyue Zhang}
\IEEEauthorblockA{Max Planck Institute for Informatics \\ Saarland Informatics Campus}
\and
\IEEEauthorblockN{Xiatian Zhu}
\IEEEauthorblockA{Samsung AI Cambridge}
\and
\IEEEauthorblockN{Chuan Sheng Foo}
\IEEEauthorblockA{Institute for Infocomm Research}
\and
\IEEEauthorblockN{Kui Jia}
\IEEEauthorblockA{South China University of Technology, China}
}
\maketitle
\vspace{-0.5cm}

\begin{abstract}
Semantic understanding of 3D point cloud relies on learning models with massively annotated data, which, in many cases, are expensive or difficult to collect.
This has led to an emerging research interest in
semi-supervised learning (SSL) for 3D point cloud.
It is commonly assumed in SSL that the unlabeled data are drawn from the same distribution as that of the labeled ones; This assumption, however, rarely holds true in realistic environments.
Blindly using out-of-distribution (OOD) unlabeled data could harm SSL performance.
In this work, we propose to selectively utilize unlabeled data through sample weighting, so that only conducive unlabeled data would be prioritized.
To estimate the weights, we adopt a bi-level optimization framework which iteratively optimizes a meta-objective on a held-out validation set and a task-objective on a training set.
Faced with the instability of efficient bi-level optimizers, we further propose three regularization techniques to enhance the training stability.
Extensive experiments on 3D point cloud classification and segmentation tasks verify the effectiveness of our proposed method.
We also demonstrate the feasibility of a more efficient training strategy.

\end{abstract}

\IEEEpeerreviewmaketitle

\section{Introduction}

Understanding 3D worlds is a fundamental and challenging problem in computer vision with a variety of critical applications for robotics, autonomous driving, etc. 3D point cloud is one of the most efficient ways to represent 3D worlds, as they can be collected from many sensors and algorithms, e.g. LiDAR and SLAM. The state-of-the-art approaches towards 3D point cloud understanding require access to a large amount of labeled data to train MLPs~\cite{qi2017pointnet}, graph convolution networks~\cite{wang2019dynamic}, or transformers~\cite{zhao2020point}. Acquiring the annotations for 3D point cloud data is nevertheless expensive due to extensive interaction with 2D projections in multiple viewpoints. Alternative to developing deeper and more complicated backbone networks, recent works explore unlabelled data to improve the performance of 3D point cloud models when labeled data is scarce~\cite{xu2020weakly,zhao2020sess,2020Label,2020PointContrast}, a.k.a. semi-supervised learning (SSL). Among these works, consistency-based SSL is proven to be particularly effective~\cite{xu2020weakly, zhao2020sess}. 
In general, it adopts a self-training paradigm by maintaining two  networks and use the pseudo-labels predicted on unlabeled data to supervise a trainable network.

\begin{figure}[!t]
\centering   
\includegraphics[width=0.9\linewidth]{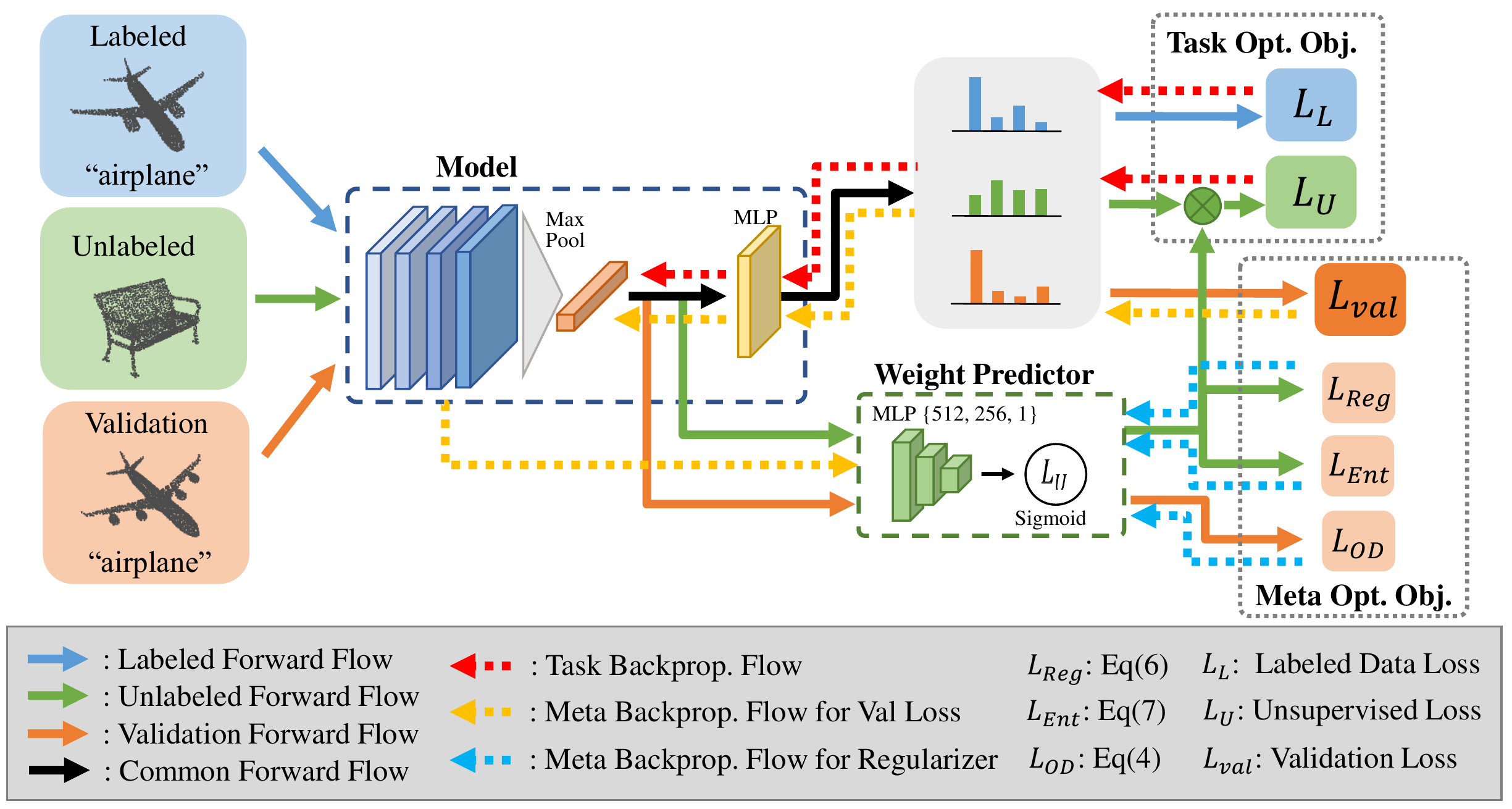}
\vspace{-0.4cm}
\caption{Overview of our Regularized Bi-level Optimization  architecture for open-set semi-supervised learning for 3D point cloud.}
\label{Schematic Diagram}
\vspace{-0.5cm}
\end{figure}

These SSL methods often assume that unlabeled data are all relevant and conducive to the learning task. However this is often not true as unlabeled data may be collected from a different domain from the labeled data likely with different data distributions and class spaces. 
Moreover, unlabeled data may not be carefully curated and are heavily contaminated by noise. For instance, regions cropped out from large-scale real scans could contain both foreground object and background, along with noise introduced by reconstruction algorithms.
Therefore, an increasingly important problem in semi-supervised 3D point cloud learning is to learn with potentially non-conducive unlabelled data, a.k.a. out-of-distribution data (OOD)~\cite{chen2020semi,guo2020safe}. SSL under such a setting is commonly referred to as open-set semi-supervised learning~\cite{yu2020multi,saito2021openmatch}.




The recent attempts to address open-set SSL can be classified into two categories. The first genre approaches from some heuristic criterion~\cite{chen2020semi,yu2020multi,saito2021openmatch}, e.g. the consistency of predictions upon multiple augmentations or training an outlier detector. A heuristic threshold is typically required to prune out outlier/out-of-distribution samples from the unlabeled data pool. Another line of research focuses on developing a principled objective to optimize the per-sample weights for unlabeled data~\cite{guo2020safe,ren2020not}, dubbed as learning based approaches. These approaches formulate open-set SSL as a bi-level optimization problem. The weights on unlabeled data~\cite{ren2020not} or a parameterized weight predictor network~\cite{guo2020safe} are regarded as hyperparameters. 
By iteratively optimizing the meta-objective, a loss defined on a separate labeled dataset, and task-objective, a loss defined on classification or segmentation, one is able to train a model with uneven weights applied to unlabeled data. A lower weight often indicates an OOD unlabeled sample. 

Despite offering a more principled framework, there are several shortcomings with existing learning based approaches. Firstly, in the bi-level optimization formulation of \cite{guo2020safe} the meta-objective and the task objective share the same labeled training data.
Consequently, a trivial solution exists with this formulation, i.e. all unlabeled data's weights equal to 0, and the meta-gradient will be constantly 0.
An alternative bi-level approach~\cite{ren2020not} treats per-sample weights as the hyperparameters to optimize and the meta-objective is defined on a separate validation set, thus avoiding the trivial solution. 
However, as the weights for all unlabeled data must be updated in every training iteration, this approach restricts itself to conducting bi-level optimization on the full dataset. When more unlabeled data becomes available, the expensive bi-level optimization procedure must be carried out again. Finally, in a gradient-based bi-level optimization, the updating of hyperparameters depends on the meta-gradient. This gradient is usually computed by approximation algorithms~\cite{ren2020not,liu2018darts,lorraine2020optimizing} and is subject to  approximation noise. Blindly using the hypergradients with off-the-shelf gradient optimizer, e.g. SGD or Adam~\cite{kingma2014adam}, will lead to  training instability.

To overcome the above disadvantages of existing bi-level optimization based approaches, we propose an open-set SSL framework with REgularized Bi-level Optimization (ReBO). Unlike existing alternatives, ReBO defines the meta-objective on a held-out validation set, thus getting rid of trivial solution. ReBO further employs a MLP as the weight predictor to obtain per-sample weights on unlabeled data. 
Importantly, this weight predictor can be pretrained on a small training subset and then transferred to the full dataset without change, resulting in more efficient model training. 
Besides, ReBO is complemented with three meta-gradient regularization techniques, including outlier detection, parameter moving averaging and entropy regularization. Extensive results validate the advantage of introduced regularizations.





We summarize the contributions of this work as below.
\begin{itemize}
  \item We formulate open-set semi-supervised learning as a bi-level optimization problem. To avoid trivial solution, meta-objective is defined on a held-out validation.
  \item A weight predictor network is introduced to estimate per-sample weights for unlabeled data. Critically, it can be trained on a subset of dataset efficiently before applied to full-scale open-set SSL without the need for further optimization.
  \item To address the instability issue of bi-level optimization,
  we introduce three regularization terms to further stabilize meta optimization loop.
\end{itemize}

\section{Related Work}
\noindent\textbf{3D Point Cloud Deep Learning.}
{3D Point cloud represents 3D object surface structure via a collection of 3D points. Common tasks on 3D Point cloud learning include classification~\cite{chang2015shapenet, wu20153d}, segmentation~\cite{mo2019partnet}. Since the emergence of deep learning based 3D point cloud understanding, PointNet \cite{qi2017pointnet} and PointNet++ \cite{qi2017pointnet++}, there has been a surge in the development of deep learning backbones, e.g.  DGCNN \cite{wang2019dynamic}, PointCNN \cite{li2018pointcnn}, CurveNet \cite{xiang2021walk}, and transformer~\cite{zhao2020point}. However,  recent success is built upon the access to large amount of labeled 3D point cloud data. Annotating 3D data is hard and how to alleviate the dependence on labeled data is becoming increasingly important.
}

\noindent\textbf{Semi-Supervised Learning.}
{
Semi-supervised learning aims to exploit large amount of unlabeled data to improve learning performance. Consistency-based SSL has demonstrated superior performance partially due to explicitly exploiting pseudo labels on unlabeled data~\cite{rasmus2015semi, tarvainen2017mean,sohn2020fixmatch}. 
The state-of-the-art SSL method~\cite{sohn2020fixmatch} generates pseudo-labels for weakly augmented samples with confident predictions, and then match them with the prediction of the strongly augmented ones. 
Inspired by the success of SSL, adaptation to 3D point cloud learning was introduced by \cite{xu2020weakly,zhao2020sess,cheng2021sspc} with successful demonstration on semantic segmentaion and object detection. 
The SSL approaches treats all unlabeled data equally while some unlabeled data could be harmful to the target task. As a result, open-set SSL emerges as a way to selectively exploit unlabeled data.}


\noindent\textbf{Open-Set Semi-Supervised Learning.}{
Open-set semi-supervised learning~\cite{chen2020semi,yu2020multi,guo2020safe,park2021opencos,saito2021openmatch} refers to the challenging case where unlabeled data may contain harmful data, sometimes referred to as out-of-distribution data (OOD).
Open-set SSL can be roughly classified into two genres. The heuristic OOD detection approaches often explicitly detect unlabeled OOD samples, through training an OOD detector~\cite{yu2020multi,saito2021openmatch} or calculating consistency of pseudo-label predictions~\cite{chen2020semi,luo2021consistency,park2021opencos}. Some heuristic thresholding is required to distinguish outliers from inliers. Alternative to the heuristic approaches, a learning-based paradigm defines a meta-objective, often as the loss on a separate labeled dataset~\cite{guo2020safe,ren2020not}, where open-set SSL is formulated as learning per-sample weight on unlabeled data such that the model trained with weighted loss minimizes the meta-objective. The whole problem can be formulated as a bi-level optimization problem. In this work, we address the unmet challenges in existing learning-based approaches and consistently improved on open-set semi-supervised 3D point cloud classification and segmentation tasks. }

\section{Methodology}
\vspace{-0.1cm}
In this section, we first briefly review semi-supervised learning and then propose an open-set semi-supervised learning framework by weighting unlabeled loss. Then we introduce the bi-level optimization procedure with newly proposed regularization terms.

\subsection{Semi-Supervised Learning}
\vspace{-0.1cm}
We first provide an overview of semi-supervised learning. Specifically, we denote the labeled training dataset as $\set{D}_{tr}^{l}=\{\matr{X}_i,\matr{y}_i\}_{i=1\cdots N_{tr}^l}$, the unlabeled training dataset as $\set{D}_{tr}^u=\{\matr{X}_j\}_{j=1\cdots N_{tr}^u}$ and the validation dataset as $\set{D}_{val}=\{\matr{X}_k,\matr{Y}_k\}_{k=1\cdots N_{val}}$. Consistency based SSL often optimizes a loss function consisting of two terms, one defined on labeled data $\mathcal{L}_{l}$ and another on unlabeled data $\mathcal{L}_{u}$. 
For classification and segmentation tasks, labeled loss is often instantiated as cross-entropy loss and unlabeled loss is instantiated as a variant of consistency between posteriors.
\vspace{-0.1cm}

Existing SSL approaches make the assumption that all unlabeled data are helpful and consistency loss is optimized on all available unlabeled data. However, this may not be true under an open-set SSL scenario where OOD unlabeled samples may exist. 
To tackle this issue, we apply non-uniform weights to each unlabeled sample by assuming that harmful samples should be given lower weights. The open-set SSL training loss $\mathcal{L}_{tr}$ now writes in Eq.~(\ref{eq:openssl}) where $\lambda(\matr{X}_j;\Phi)$ is a function determining a per-sample weight applied to unlabeled samples.
\vspace{-0.1cm}
\begin{equation}\label{eq:openssl}
\resizebox{0.9\linewidth}{!}{$\min\limits_{\Theta} \frac{1}{N_{tr}^l}\sum\limits_{\matr{X}_i,\matr{y}_i\in\set{D}_{tr}^l} \mathcal{L}_{l}(\matr{X}_i,\vect{y}_i) + \frac{1}{N_{tr}^u}\sum\limits_{\matr{X}_j\in\set{D}_{tr}^u} \lambda(\matr{X}_j;\Phi)\mathcal{L}_{u}(\matr{X}_j)$}
\vspace{-0.1cm}
\end{equation}
To obtain the per-sample weight, we use a neural network to instantiate the function $\lambda(\matr{X}_j;\Phi)$. Therefore a hyperparameter, $\Phi$, is introduced in this formulation. Alternative to this design, \cite{ren2020not} directly treats per-sample weights as hyperparameters. We argue that predicting weights through a network is more flexible and light-weight in that we can learn such a network on a subset of data and then transfer to the full-scale SSL. 

\subsection{Bi-Level Optimization}\label{sect:DARTS}
\vspace{-0.1cm}
Contrary to regular neural network training where hyperparameters are often determined by cross-validation, the open-set SSL formulate requires hyperparameters, $\Phi$, to be simultaneously updated with model parameters, $\Theta$. As stated previously, we wish to adjust $\Phi$ such that the performance on a held-out validation set is maximized, therefore, the following bi-level optimization formulation is adopted.
\vspace{-0.1cm}
\begin{equation}\label{eq:bilevel}
{\begin{split}
  &\min_{\Phi} \frac{1}{N_{val}}\sum_{\matr{X}_k,\matr{Y}_k\in\set{D}_{val}} \mathcal{L}_{l}(\matr{X}_k,\vect{Y}_k;\Theta^*),\\ & s.t.\quad \Theta^*=\arg\min_{\Theta} \frac{1}{N_{tr}^l}\sum_{\matr{X}_i,\matr{y}_i\in\set{D}_{tr}^l} \mathcal{L}_{l}(\matr{X}_i,\vect{y}_i) \\&+ \frac{1}{N_{tr}^u}\sum_{\matr{X}_j\in\set{D}_{tr}^u} \lambda(\matr{X}_j;\Phi)\mathcal{L}_{u}(\matr{X}_j)
\end{split}
}
\vspace{-0.2cm}
\end{equation}

The problem in the first row is often referred to as meta-objective optimization and we denote the loss defined on the validation set as the validation loss $\mathcal{L}_{val}$. The nested problem in the second and third row is referred to as task-objective optimization.
Solving the bi-level problem in Eq.~(\ref{eq:bilevel}) is non-trivial. For gradient based optimization one has to compute the meta-gradient ${\nabla_{\Phi}\mathcal{L}_{val}}$. Since validation loss is not an explicit function of $\Phi$ we opt for a single-step unrolling algorithm~\cite{liu2018darts} to approximate the gradient. In specific, we first use the chain rule to expand the meta-gradient as,
\begin{equation}\label{eq:metagradient}
  \nabla_{\Phi}\mathcal{L}_{val}=\frac{\partial\mathcal{L}_{val}(\Theta^*)}{\partial\Phi}=\frac{\partial\mathcal{L}_{val}(\Theta^*)}{\partial\Theta^*}\frac{\partial\Theta^*}{\partial\Phi}
\end{equation}

For any differential validation loss, e.g. cross-entropy loss, the first term $\frac{\partial\mathcal{L}_{val}(\Theta^*)}{\partial\Theta^*}$ is computed analytically. Calculating the second term involves differentiation over a trajectory of $\Theta$ if gradient descent style updating is adopted for task optimization. For efficient optimization, we adopt the one-step approximation proposed in \cite{liu2018darts} $\Theta^*=\Theta-\alpha\nabla_{\Theta}\mathcal{L}_{tr}$, where $\alpha$ is set equal to the learning rate for task optimization. The meta-gradient in Eq.~(\ref{eq:metagradient}) now writes $\nabla_{\Phi}\mathcal{L}_{val}=-\alpha\nabla_{\Phi,\Theta}^2\mathcal{L}_{tr}(\Theta,\Phi)\nabla_{\Theta^*}\mathcal{L}_{val}(\Theta^*,\Phi)$. The Hessian-vector multiplication can be efficiently approximated by finite difference approximation as below where $\Theta^\pm=\Theta\pm \epsilon\nabla_{\Theta^*}\mathcal{L}_{val}({\Theta^*,\Phi})$
and $\epsilon=1e^{-2}/\left \| \nabla_{\Theta^*}\mathcal{L}_{val}({\Theta^*,\Phi}) \right \|_2$.
\begin{equation}
  \nabla_{\Phi}\mathcal{L}_{val}\approx-\alpha\frac{\nabla_{\Phi}\mathcal{L}_{tr}(\Theta^+,\Phi)-\nabla_{\Phi}\mathcal{L}_{tr}(\Theta^-,\Phi)}{2\epsilon}
\end{equation}


\subsection{Regularizing Meta-Gradient}\label{sect:Regularization}

Learning weight predictor by minimizing validation loss is challenging as the meta-gradient, Eq.~(\ref{eq:metagradient}) is often noisy due to the multiple approximations adopted. To improve the training stability we propose to add further regularizations to help optimize meta-objective. 

\noindent\textbf{Regularizing by Outlier Detection.}
We first propose to incorporate a regularization inspired by outlier detection. 
As with the heuristic open-set SSL approaches~\cite{yu2020multi}, discovering out-of-distribution is often treated in a similar fashion to outlier detection~\cite{ruff2018deep}. Under this assumption, the majority of training samples are regarded as normal samples and thus the encoder network is forced to embed features in a cluster, a.k.a. one-class classification. Inspired by this design, given a clean (in-distribution) validation set, we propose to add a classification branch such that all validation set should have a uniform weight. More specifically, the following cross-entropy loss is adopted. 
\vspace{-0.1cm}
\begin{equation}\label{eq:LOD}
  \mathcal{L}_{OD}=\sum_{\matr{X}_k\in\set{D}_{val}}-\log\lambda(\matr{X}_k;\Phi)
\end{equation}
We further notice the meta-gradient computed from approximation algorithms may not always point to the correct direction that reduces the meta-objective. 
As a result, optimizing hyperparameters with meta-gradients alone is empirically unstable and sometimes may not converge at all.
Therefore, we propose a regularization to smooth out the noisy meta-gradients.

\noindent\textbf{Disappearing Tikhonov Regularization.}
We wish to regularize the weight predictor's output to be smooth over iterations. Therefore, we first notice that the Disappearing Tikhonov regularization (DTR) as below helps smooth parameters temporally, where ${w}_{t-1}$ is the weight predicted in the previous iteration. 
\vspace{-0.2cm}
\begin{equation}
  \mathcal{L}_{reg}=||\lambda_t(\matr{X}_j;\Phi)-{w}_{t-1}||_2^2
\end{equation}

\noindent\textbf{Moving Averaging Regularization.}
Despite DTR regulerize weights temporally, it only considers the weight one-step back. To enforce stronger temporally smoothness, we propose to maintain a moving average of unlabeled sample weights $\tilde{w}_t=\beta \tilde{w}_{t-1} + (1-\beta)w_t$ as target for regularization. This results in the Moving Averaging regularization.
\vspace{-0.1cm}
\begin{equation}\label{eq:Reg-MA}
  \mathcal{L}_{reg}=||w_t-\tilde{w}_{t-1}||_2^2=||\lambda_t(\matr{X}_j;\Phi)-\tilde{w}_{t-1}||_2^2
\end{equation}






\noindent\textbf{Entropy Regularization}:
Since the per-sample weights are continuous, it is not guaranteed the harmful samples will receive weights that are low enough to eliminate the negative impact. Therefore, the weight predictor is further forced to produce output close to binary distribution. We add an entropy regularization term for this purpose, defined as below. 
\vspace{-0.2cm}
\begin{equation}\label{eq:Entropy}
\begin{split}
  L_{ent}=&-\frac{1}{|\set{D}_{tr}^u|}\sum_{\matr{X}_j\in\set{D}_{tr}^u }\lambda(\matr{X}_j;\Phi)\log\lambda(\matr{X}_j;\Phi)\\
  &+(1-\lambda(\matr{X}_j;\Phi))\log \lambda(\matr{X}_j;\Phi)
\end{split}
\end{equation}

Eventually, we combine all regularization terms and formulate the final meta optimization objective as,
\vspace{-0.1cm}
\begin{equation}\label{eq:finalgradient}
  \min_{\Phi} \mathcal{L}_{val} + \gamma \mathcal{L}_{reg} + \xi \mathcal{L}_{ent} + \eta \mathcal{L}_{OD}
\end{equation}

All regularizers have analytic gradient w.r.t. $\Phi$ and can be linearly combined with approximated validation gradient as the final meta-gradient.






\subsection{Training Strategy} \label{sect:Training Strategy}
{Bi-level optimization is expensive when unlabeled data is huge. To avoid the expensive training, we propose two efficient warm-up training strategies for open-set SSL. In specific, we opt for a less computational demanding paradigm by pre-training the backbone and predictor network on labeled data and a subset of unlabeled data. After convergence, we infer and fix the weights for all unlabeled data, and later transfer them to retrain a new backbone from scratch, termed as \textbf{Transfer} approach. It's worth noting that only backbone is updated during retraining, thus it saves substantial time for bi-level optimizing weight predictor. Alternatively, we take the pre-trained parameters to initialize a backbone and a predictor, and then fine-tune on full data for just a few epochs, termed as \textbf{Fine-tune} approach. Although both warm-up methods can significantly save training time, they bring different enlightenment. Weights transferring implies that the predicted weights in our methods could be transferred to other backbones, as a fixed measure of data fitness. While fine-tuning suggests a general bi-level training paradigm for large-scale open-set SSL, i.e. dividing a subset data for pilot training. }

\section{Experiment}
\subsection{Dataset}
{We evaluate our methods on three datasets including ModelNet40\cite{wu20153d}, ShapeNet Part\cite{mo2019partnet} and ObjectNN\cite{uy2019revisiting}.}
{\textbf{ModelNet40} is a shape classification benchmark dataset consisting of 12,311 CAD shape models from 40 categories, split into 9,843 for training and 2,468 for testing. For each shape, we uniformly sample 1024 points on its CAD mesh and normalize them into a unit sphere.}
{\textbf{ShapeNet Part} contains 16,881 shapes from 16 categories with 50 part categories, split into 12,137 for training, 1,870 for validating and 2,874 for testing. For each shape model, 2048 points are uniformly sampled from surface mesh.}
{\textbf{ObjectNN} is a real scan object dataset, where 2,902 objects in 15 categories are scanned from real scene ScanNet\cite{dai2017scannet} and SceneNN\cite{hua2016scenenn} and each object is uniformly sampled with 2048 points.}

\noindent\textbf{Data Split}:{ Throughout the training we keep the amount of all labeled data to be fixed. At every new epoch, the validation set is randomly drawn from the whole labeled dataset and the rest are used as labeled training data with a 1:1 ratio. As a result, all labeled data could be used for updating task network throughout the whole training episode.}

\subsection{Configuration}
\noindent\textbf{Encoder Network.}
{{We choose DGCNN~\cite{wang2019dynamic} as the default backbone for point cloud classification and part segmentation. In addition to DGCNN, we further evaluate with CurveNet~\cite{xiang2021walk} for transfer experiments.
}

\noindent\textbf{Weight Predictor.}{ Our weight predictor is composed of a three-layer MLP with batch normalization and relu activation following each linear layer. The output layer is normalized to between 0 and 1 by a sigmoid function. For  classification network, weight predictor takes the backbone's global feature (after maxpooling) as input. In part segmentation task, we append maxpooled feature to point-wise feature as input to weight predictor.}


\noindent\textbf{Warm-up.}{
Randomly-initialized weight predictor is very unstable, resulting in great fluctuations of predicted weights in the beginning. Inspired by Multi-OS \cite{yu2020multi}, we pre-train backbone and weight predictor for 30 epochs to obtain good initial weights. Specifically, we pre-train backbone with labeled loss only, and at the meantime, we guide weight predictor to output zero-weights for all unlabeled samples while one-weights for validation samples by a cross entropy loss. 
}}

\noindent\textbf{Hyper-Parameters.}
{{During backbone training, we adopt  cross entropy loss as labeled loss and FixMatch \cite{sohn2020fixmatch} consistency loss as unlabeled loss. In addition, loss combination in Eq.~\ref{eq:finalgradient} is used to optimize weight predictor. Here we set the rate $\beta$ acting in Eq.~\ref{eq:Reg-MA} to 0.5, and the constraint hyper-parameters $\gamma$, $\eta$ in Eq.~\ref{eq:finalgradient} are set to 0.1, 0.01 empirically. As a strong entropy regularization may prevent weight predictor from optimizing properly at the beginning of training, we gradually increase $\xi$ over the training episode. We defer the details of $\xi$ updating to the supplementary. 
}}

\noindent\textbf{Competing Methods.}{
Our baseline is conducted with DGCNN backbone  trained by FixMatch semi-supervised loss with all unlabeled weights fixed to 1. We compare our method with 4 most relevant and up-to-date methods, DS3L\cite{guo2020safe}, Multi-OS\cite{yu2020multi}, LTWA\cite{ren2020not} and OP-Match \cite{saito2021openmatch}. We migrate them to our point cloud setting and combine them with the same FixMatch SSL. Because LTWA needs to calculate meta-objective on a separate validation set, we adopt the same data split strategy to dynamically divide training and validating set in each epoch. }

\subsection{SSL with Out-of-Distribution Data}
{We create an open-set ssl setting by incorporating different OOD data into the unlabeled dataset. Unlike \cite{luo2021consistency} and \cite{yu2020multi} which mix the unlabeled target data with an extra OOD dataset as the whole unlabeled set, we consider that there could be more than one type of OOD data present in unlabeled set. It means some OOD samples may be extremely harmful for backbone learning while others may be less harmful or even useful. Therefore, we define a weak and a strong OOD  sets. We choose S3DIS\cite{20163D} to create OOD set for classification and part segmentation experiments. S3DIS is a large indoor scene point cloud dataset, covering 271 rooms from 6 areas. We follow PointNet~\cite{qi2017pointnet} to divide each room into 1*1 meter blocks and randomly sample 1024/2048 (classification/part segmentation) points from each blocks. 10,000 randomly selected blocks are treated as weak OOD data ($\set{W}$) because some blocks may just crop out objects, like chair, table, etc., that appears in the target dataset. In addition, we select another 10,000 samples and augment them by random rotating with a maximum amplitude of $90^{\circ}$ and Gaussian jittering $\mathcal{N}(0,1)$, as strong OOD set($\set{S}$). This augmentation will destroy even the manifold of good objects, thus be deemed strong OOD.}

\subsubsection{Open-set SSL Classification}
{We first evaluate on ModelNet40 shape classification benchmark. For training, we randomly sampled $\{100, 400, 1000\}$ samples as labeled set $\set{L}$ and the rest as in-distribution unlabeled data $\set{U}$. In addition, we append S3DIS $\set{W}$ and $\set{S}$ as OOD unlabeled set. We report overall classification accuracy with different methods in Tab.~\ref{Classification}. We first show four baseline results under different combinations of data in the top section of the table. It can be seen that conducting semi-supervised learning with in-distritbuion unlabeled data can significantly improve backbone accuracy(from 55.1$\%$ to 62.1$\%$ @ 100 labeled). However, with weak OOD samples in unlabeled pool we observe a clear drop of performance (from 62.1$\%$ to 60.2$\%$ @ 100 labeled). When strong OOD samples are included $\{\mathcal{L},\mathcal{U},\mathcal{W},\mathcal{S}\}$, a more significant loss of peformance is observed (from 60.2$\%$ to 58.8$\%$ @ 100 labeled). 

We then evaluate all competing methods and our proposed ReBO method on the full dataset $\{\mathcal{L},\mathcal{U},\mathcal{W},\mathcal{S}\}$. We observe all open-set SSL methods improves over the baseline thanks to the ability to detect and downweight OOD samples. In particular, ReBO achieves the highest classification results under all three amounts of labeled data, even higher than the baseline with completely clean data ($\set{L}$ + $\set{U}$). This suggests that ReBO is able to discover some useful examples from the weak $\mathcal{W}$ and strong $\mathcal{S}$ OOD samples and improve performance with these data as additional unlabeled data.
}
\subsubsection{Open-set SSL Segmentation}
{We conduct part-segmentation experiments on ShapeNet with $\{20, 100, 400\}$ random labeled samples $\set{L}$ and the other as in-distribution unlabeled data $\set{U}$. We also add S3DIS $\set{W}$ and $\set{S}$ in all unlabeled data. The mean IoU over all samples are showed in Tab.~\ref{Part Seg}. First, adding additional weak OOD dataset does not necessarily harm the baseline performance. We hypothesize the reason being segmentation focuses more on local parts and geometry and the weak OOD data are cropped out from real scanned data thus carrying more diverse local geometry that benefits the segmentation task. Moreover, compared with other methods, our method again performs the best in all three unlabeled data settings. At the higher labeling regimes (100, 400 labeled) ReBO again beats baselines without strong OOD samples.
}

\subsection{Open-set SSL on Real Scanned Data} \label{sect:Real Scan Experiments}
{Compared with synthetic data, real scanned data is more meaningful and challenging. We now further validate the performance of ReBO on real scanned ObjectNN dataset~\cite{uy2019revisiting}. We assume that objects scanned from real scenes are not perfectly cropped and are subject to background noise. To simulate this secanrio, we perturb the object bounding boxes provided in ScanNet and SceneNN, and then crop out OOD point cloud samples from perturbed boxes. In the experiments, we collect 2,000 noisy OOD samples denoted as $\set{O}$ to complement the labeled data $\set{L}$ and unlabeled data $\set{U}$. The final results in Tab.~\ref{Real Scanned Experiments} demonstrate that open-set methods generalize to practical point cloud situation, and our method still performs the best among all competing methods.
}

\begin{table}[!t]
\caption{Accuracy ($\%$) for different methods with different numbers of labeled samples on ModelNet40 Classification.}
\vspace{-0.3cm}
\label{Classification}
\centering
\scalebox{0.75}{
\begin{tabular}{c c c c c}
  \toprule
  Data & Methods & 100 labeled & 400 labeled & 1000 labeled \\	
  \hline
  $\{\set{L}\}$ & Baseline & $55.1$ & $78.4$ & $83.3$\\
  $\{\set{L},\set{U}\}$ & Baseline & $62.1$ & $80.5$ & $84.7$ \\
  $\{\set{L},\set{U},\set{W}\}$ & Baseline & $60.2$ & $79.4$ & $84.0$\\
  \hline
  \multirow{6}{*} {$\{\set{L},\set{U},\set{W},\set{S}\}$} & Baseline & $58.8$ & $78.9$ & $83.7$ \\
  & DS3L \cite{guo2020safe} & $59.6$ & $79.3$ & $84.4$ \\
  & Multi-OS \cite{yu2020multi} & $59.2$ & $79.1$ & $83.6$ \\
  & LTWA \cite{ren2020not} & $60.4$ & $79.5$ & $84.9$ \\
  & OP-Match \cite{saito2021openmatch} & $61.1$ & $80.0$ & $84.6$ \\
  \cline{2-5}
  & ReBO & \bm{$63.4$} & \bm{$80.9$} & \bm{$85.5$}  \\
  \bottomrule
\end{tabular}
}
\vspace{-0.30cm}
\end{table}

\begin{table}[!t]
\caption{mIoU($\%$) for different methods with different numbers of labeled samples on ShapeNet Part Segmentation.}
\vspace{-0.3cm}
\label{Part Seg}
\centering
\scalebox{0.75}{
\begin{tabular}{c c c c c}
  \toprule
  Data & Methods & 20 labeled & 100 labeled & 400 labeled \\	
  \hline
    $\{\set{L}\}$ &	Baseline & $54.0$ & $72.6$ & $78.1$ \\
  $\{\set{L},\set{U}\}$ & Baseline & $55.5$ & $73.9$ & $78.8$ \\
  $\{\set{L},\set{U},\set{W}\}$ & Baseline & $56.3$ & $74.6$ & $78.0$\\
  \cline{1-5}
  \multirow{6}{*} {$\{\set{L},\set{U},\set{W},\set{S}\}$} & Baseline & $55.3$ & $74.0$ & $77.7$\\	
  & DS3L \cite{guo2020safe} & $55.4$ & $73.6$ & $76.4$ \\
  & Multi-OS \cite{yu2020multi} & $56.0$ & $74.4$ & $78.1$ \\
  & LTWA \cite{ren2020not} & $54.3$ & $74.1$ & $78.7$\\
  & OP-Match \cite{saito2021openmatch} & $55.8$ & $74.7$ & $78.4$ \\
  \cline{2-5}
  & ReBO & \bm{$56.1$} & \bm{$75.5$} & \bm{$79.6$}\\
  \bottomrule
\end{tabular}
}
\vspace{-0.3cm}
\end{table}

\begin{table}[!t]
\caption{Classification accuracy ($\%$) on real-world ObjectNN Experiments.}
\vspace{-0.3cm}
\label{Real Scanned Experiments}
\centering
\begin{tabular}{c c c c}
  \toprule
  Data & Methods & $5\%$ & $20\%$  \\	
  \hline
    $\{\set{L}\}$ &	Baseline & $37.5$ & $66.1$ \\
  $\{\set{L},\set{U}\}$ &	Baseline & $40.4$ & $67.5$ \\
  \hline
  \multirow{6}{*} {$\{\set{L},\set{U},\set{O}\}$} & Baseline & $38.2$ & $66.4$  \\
  & DS3L \cite{guo2020safe} & $38.6$ & $65.0$ \\
  & Multi-OS \cite{yu2020multi} & $39.3$ & $65.1$ \\
  & LTWA \cite{ren2020not} & $39.0$ & $67.7$ \\
  & OP-Match \cite{saito2021openmatch} & $39.5$ & $67.1$\\
  \cline{2-4}
  & ReBO & \bm{$40.6$} & \bm{$68.5$} \\
  \bottomrule
\end{tabular}
\vspace{-0.5cm}
\end{table}

\subsection{Additional Evaluation}
\noindent\textbf{Continual Learning on Unseen Data}\label{sect:Unseen Generalization}
{In a realistic open-set SSL environment the unlabeled data could continuously expand when more data is collected. Given the expensive computation cost of bi-level optimization, training from scratch when more unlabeled data becomes available is a very inefficient practice. Therefore, we consider conitnual learning on unseen data task in this section to simulate the ever expanding unlabeled dataset.
Formally, we consider ShapeNet as an additional unseen dataset in our ModelNet40 classification experiments. When the backbone and weight predictor have been trained on ModelNet40($\set{L}$ + $\set{U}$) and S3DIS($\set{W}$ + $\set{S}$), we could estimate ShapeNet samples' weights directly without additional training. The weights on ShapeNet are fixed for continual training(50 epochs) of backbone. In addition, we could fine-tune our backbone and predictor for 50 epochs with additional unseen ShapeNet data. We explore these two options and present the results in Tab.~\ref{Generalization to Unseen Data}. It can be found that the model does benefit from additional unseen data with 1.9$\%$ @ 100 labeled and 0.6$\%$ @ 400 labeled improvements even without unlabeled data weighting. But further improvement is observed if we estimate and fix the unlabeled data weights (Est $\&$ Fix) , suggesting the benefits of selectively using unlabeled data. Finally, the best results is obtained by doing further fine-tuning (Fine-tune) with additional unseen data. This cheap fine-tuning approach turns out to be more effective.} 

\noindent\textbf{Comparing with SOTA SSL}
{In this section, we compare with existing label-efficient learning methods on ShapeNet part segmentatoin benchmark.
Similarly, we divide ShapeNet into $\set{L}$ and $\set{U}$. Moreover, we could have ModelNet40 as additional unlabeled data $\set{M}$. For the competing methods, ScanNet $\mathcal{N}$ and ShapeNetCore~$\mathcal{C}$ were respectively used as unlabeled data. In Tab.~\ref{SOTA SSL ShapeNet PartSeg}, we compare our method with SO-Net \cite{2018SO}, PointCapsNet\cite{zhao20193d}, JointSSL\cite{alliegro2021joint}, Multi-task \cite{2019Unsupervised}, PCont \cite{2020PointContrast}, ACD \cite{2020Label} under $\{1\%, 5\%\}$ labeled samples($\set{L}$). We demonstrate that with the help of unlabeled data $\set{U}$ and additional ModelNet40 $\set{M}$, we can achieve much better results than the competing methods.}

\noindent\textbf{Additional Backbone}
{
To prove that our weight predictor is also applicable to the latest task networks, we swap our backbone with CurveNet \cite{xiang2021walk}, a point cloud shape analysis network proposed to obtain local information by curve walking. We first repeat ModelNet40 classification experiments and present results in Tab.~\ref{CurveNet Classification}. It shows that our weight predictor method still outperforms all other methods with state-of-the-art backbone network, reaching 69.1$\%$ / 81.4$\%$ accuracy under 100 / 400 labeled samples setting. We further explore the universality of weights across different backbones. We fix and transfer the weights estimated by DGCNN to CurveNet SSL training. The result (weight transfer) proves that CurveNet could work with the transferred weights and it is only slightly behind training from scratch. This indicates estimating unlabeled data weights with a light-weight model is a feasible way to save computation cost.
}

\noindent\textbf{Computation Complexity}
In Tab.~\ref{Comparison of computations}, we compare the number of all parameters and the training time cost with DGCNN bacckbone on ModelNet40 with 100 labeled samples. Compared with outlier detection methods~(Multi-OS, OP-Match), the weight optimization methods~(DS3L, LTWA, ReBO without finetune) is more computationally expensive. Our proposed fine-tune strategy reduced training time from 231 hours to 81 hours. Overall, compared with the baseline method we achieved 4.6$\%$ improvement in accuracy with only $60\%$ more training time.\\

\begin{table}[!t]
\caption{Transferring to Unseen Data. Uniform Unseen Weights means all weights in unseen data are set to 1. Est \& Fix and Fine-tune are two ways of our continual training.}
\vspace{-0.3cm}
\label{Generalization to Unseen Data}
\centering
\scalebox{1.0}{
\begin{tabular}{c c c c}
  \toprule
  Methods & 100 labeled & 400 labeled \\
  \hline
  w/o Unseen Data & $63.4$ & $80.9$ \\
  Uniform Unseen Weights & $65.3$ & $81.5$ \\
    \hline
  Est \& Fix & $66.3$ & $82.1$ \\
  Fine-tune & \bm{$67.1$} & \bm{$82.4$} \\
  \bottomrule
\end{tabular}
}
\vspace{-0.3cm}
\end{table}

\begin{table}[!t]
\caption{Part segmentation on ShapeNet with limited labeled training data. $\set{N}$, $\set{C}$ and $\set{M}$ indicate ScanNet, ShapeNetCore and ModelNet40, respectively.}
\vspace{-0.3cm}
\label{SOTA SSL ShapeNet PartSeg}
\centering
    \scalebox{0.78}{
\begin{tabular}{c c c c}
  \toprule
  Data & Methods & $1\%$ & $5\%$  \\	
  \hline
  \multirow{4}{*} {$\{\set{L}$\}} & SO-Net\cite{2018SO} & $64.0$ & $69.0$ \\
  & PointCapsNet\cite{zhao20193d} & $67.0$ & $70.0$ \\
  & JointSSL\cite{alliegro2021joint} & $71.9$ & $77.4$ \\
  & Multi-task\cite{2019Unsupervised} & $68.2$ & $77.7$ \\
  \hline
  $\{\set{L},\set{N}\}$ & PCont\cite{2020PointContrast} & $74.0$ & $79.9$ \\
  $\{\set{L},\set{C}\}$ & ACD\cite{2020Label} & $75.7$ & $79.7$\\
  \hline
  $\{\set{L},\set{M}\}$ & ReBO & $76.2$ & $80.1$ \\
  $\{\set{L},\set{U},\set{M}\}$ & ReBO & \bm{$76.9$} & \bm{$80.3$} \\
  \bottomrule
\end{tabular}}
\vspace{-0.3cm}
\end{table}

\begin{table}[!t]
\caption{Evaluation of our method conducted with CurveNet backbone on ModelNet40 classification.}
\vspace{-0.3cm}
\label{CurveNet Classification}
\centering
\scalebox{0.85}{
\begin{tabular}{c c c c}
  \toprule
  Data & Methods & 100 labeled & 400 labeled\\
  \hline
  \multirow{7}{*} {$\{\set{L},\set{U},\set{W},\set{S}\}$} & Baseline & $67.2$ & $79.9$ \\
  & DS3L \cite{guo2020safe} & $67.8$ & $79.6$ \\
  & Multi-OS \cite{yu2020multi} & $67.4$ & $80.1$ \\
  & LTWA \cite{ren2020not} & $68.0$ & $79.9$ \\
  & OP-Match \cite{saito2021openmatch} & $67.8$ & $79.7$ \\
  \cline{2-4}
  & ReBO(weight transfer) & $68.3$ & $80.7$ \\
  & ReBO(train from scratch) & \bm{$69.1$} & \bm{$81.4$} \\
  \bottomrule
\end{tabular}
}
\vspace{-0.3cm}
\end{table}

\begin{table}[!t]
\caption{Comparison of parameters and computation cost on ModelNet40 classification task with DGCNN backbone.}
\vspace{-0.3cm}
\label{Comparison of computations}
\centering
\scalebox{0.75}{
\begin{tabular}{c c c c}
  \toprule
  Methods & Parameters~(million) & Training Time~(hours) & Accuracy~($\%$)\\
  \hline
  Baseline & $\sim1.8$ & $\sim50$ & $58.8$\\
  DS3L & $\sim1.9$ & $\sim182$ & $59.6$\\
  Multi-OS & $\sim1.9$ & $\sim52$ & $59.2$\\
  LTWA & $\sim1.8$ & $\sim421$ & $60.4$\\
  OP-Match & $\sim1.8$ & $\sim57$ & $61.1$\\
  \cline{1-4}
  ReBO~(w/o finetune) & $\sim3.0$ & $\sim231.0$ & $62.1$\\
  ReBO~(w finetune) & $\sim3.0$ & $\sim81.0$~(1.6$\times$ baseline) & $63.4$\\
  \bottomrule
\end{tabular}
}
\vspace{-0.3cm}
\end{table}

\subsection{Ablation Study}
{We conduct ablation experiments on 100 labeled samples $\set{L}$ and the rest $\set{U}$ from ModelNet40 classification with $\set{W}$ and $\set{S}$ from S3DIS.}

\noindent\textbf{Regularization}
{We first validate the importance of weight predictor, entropy loss and outlier detection (OD). As is shown in Tab.~\ref{Table Ablation}, baseline ($58.8\%$) is the result of backbone directly trained on $\{\set{L},\set{U},\set{W},\set{S}\}$. Though adding a pure weight predictor improves baseline by only $0.2\%$, the later joined Entropy in Eq.~\ref{eq:Entropy} and OD in Eq.~\ref{eq:LOD} enhance classification performance by $1.4\%$ and $0.6\%$, respectively. In addition, we compare Disappearing Tikhonov Regularization(DTR) and Moving Averaging Tikhonov Regularization(MATR) in Sect.~\ref{sect:Regularization} and conclude that MATR gives better performance with a larger $1.1\%$ improvement.}

\begin{table}[!t]
\caption{Ablation study. DTR indicates Disappearing Tikhonov regularization. MATR indicates Moving Averaging Tikhonov regularization.}
\vspace{-0.3cm}
\label{Table Ablation}
\centering
\scalebox{0.9}{
\begin{tabular}{c c c c c c c}
  \toprule
     Predictor & Entropy & OD & TmpReg & Transfer & Fine-tune & Acc$(\%)$ \\
  \hline
      -    &     -     &  -        &    -    &      -   &      -    & $58.8$ \\
  $\checkmark$ &     -        &  -           &    -         &      -   &      -    & $59.0$ \\
  $\checkmark$ & $\checkmark$ &  -           &    -         &      -   &      -    & $60.4$ \\
  $\checkmark$ & $\checkmark$ & $\checkmark$ &    -         &      -   &      -    & $61.0$ \\
  $\checkmark$ & $\checkmark$ & $\checkmark$ & DTR &      -   &      -    &  $61.7$ \\
  $\checkmark$ & $\checkmark$ & $\checkmark$ & MATR &      -   &      -    & $62.1$ \\
  $\checkmark$ & $\checkmark$ & $\checkmark$ & MATR & $\checkmark$ &      -   & $62.7$ \\
  $\checkmark$ & $\checkmark$ & $\checkmark$ & MATR &      -   & $\checkmark$ & $63.4$ \\
  \bottomrule
\end{tabular}
}
\vspace{-0.5cm}
\end{table}

\noindent\textbf{Transfer and Fine-tune}
{Estimating per-sample weight with bi-level optimization is expensive, we evaluate the two approaches, Transfer and Fine-Tune, introduced in Sect.~\ref{sect:Training Strategy}. In specific, we first train both backbone and predictor on labeled data and a subset of unlabeled data ($10\%$), and then respectively evaluate Transfer and Fine-Tune strategies. As we observe from Tab.~\ref{Table Ablation}, Transfer and Fine-tune(for 50 epochs) could improve the accuracy to $62.7\%$ and $63.4\%$, respectively. It proves that pre-training is feasible for training bi-level optimization and it saves substantial computation cost by training on only $10\%$ unlabeled data.
}

\subsection{Analysis on Predicted Weights}
\noindent\textbf{Temporal Evolution of Predicted Weights}
{We analyze and visualize predicted weights on 100 labeled ModelNet40 ($\set{L},\set{U}$) classification experiment with additional S3DIS ($\set{W},\set{S}$). In Fig.~\ref{Weights Diagram}, we show the weights of $\set{U}$, $\set{W}$ and $\set{S}$ under our different modules over training epochs. The x-axis is the number of training epoch. It can be seen from the top-left plot that when no regularization constraint is applied, although predictor assigns higher weights to $\set{U}$, it fails to converge. It is because in the later stage of training, the converged classification network will receive weaker gradients from unsupervised loss, thus producing unstable meta-gradient. The top-right plot shows that entropy loss, which forces predicted weights to 0 or 1, could stabilize predictor in the later stage of training. In the bottom-left plot, applying outlier detection could effectively distinguish $\set{W}$ and $\set{S}$, as it is conducive to focusing predictor on object feature. However, it results in increasing $\set{W}$ weights in the end. The plot at the bottom-right shows the effect of the final model, which significantly smooths and stabilizes the predicted weights. In the end, predictor could reasonably distinguish $\set{U}$, $\set{W}$ and $\set{S}$, assigning higher weights to $\set{U}$ and lower weights to $\set{S}$.
}

\begin{figure}[!t]
\centering   
\includegraphics[width=0.48\textwidth]{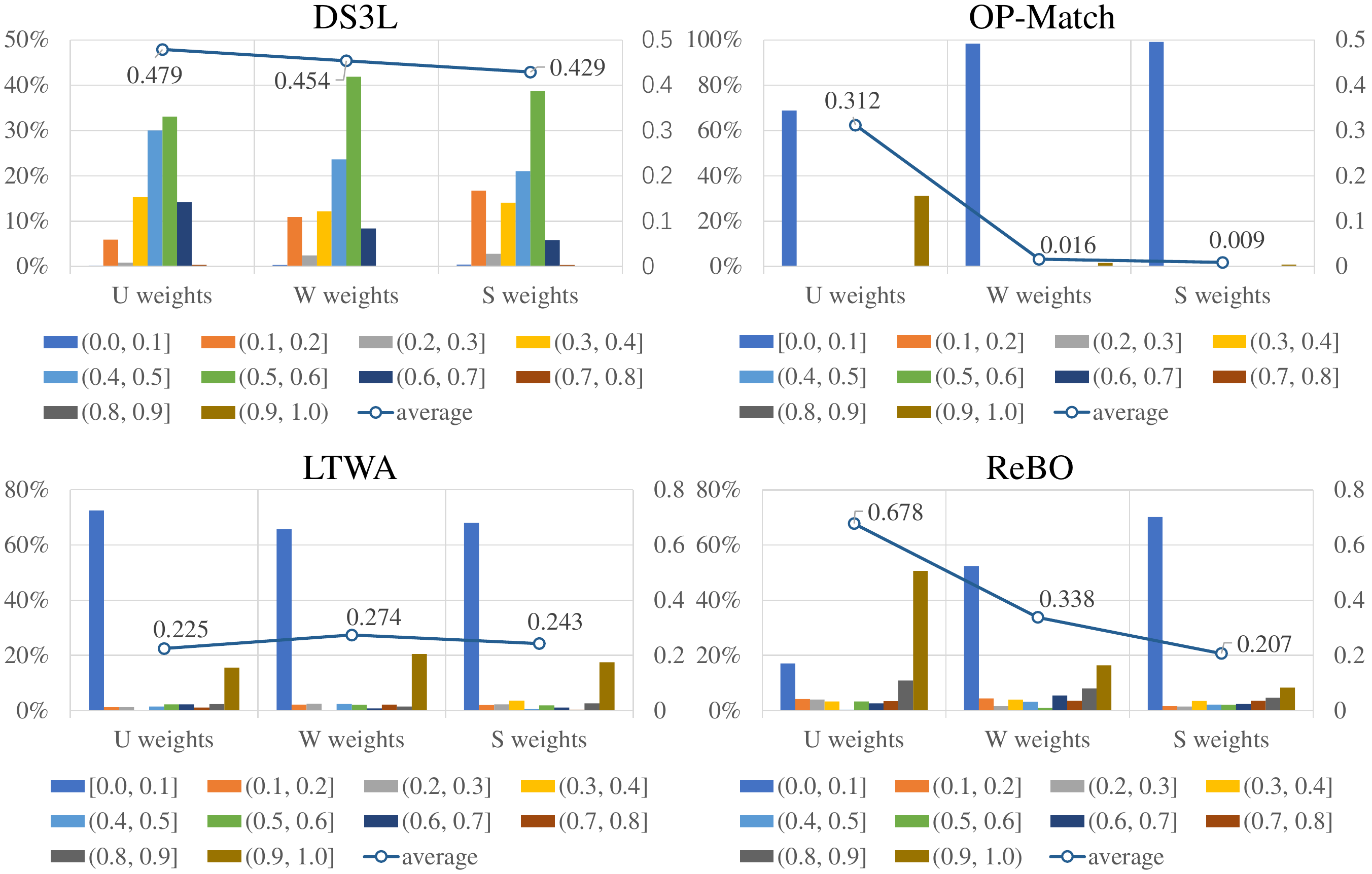}
\caption{Distribution histograms and average lines of predicted weight by different methods. The histogram percentage scale y-axis is on the left, while the average scale is on the right.}
\label{Weights Histogram}
\vspace{-0.4cm}
\end{figure}

\noindent\textbf{Distribution of Predicted Weights}{
In Fig.~\ref{Weights Histogram}, we visualize the distribution histograms of final predicted weights by different methods. In detail, we divide ten intervals evenly between 0 and 1, and plot the weights as histograms according to its proportion. For example, in the top-left figure, the orange column of $\set{U}$ weights occupies nearly 5$\%$, which means that almost 5$\%$ of $\set{U}$ weights are in the interval (0.1, 0.2]. Moreover, we present the plot of average weight, where the first point means that the average of $\set{U}$ weights is 0.479. As shown in the top-left figure, weights produced by DS3L are concentrated around 0.5, and $\set{U}$, $\set{W}$ and $\set{S}$ have similar average values. Multi-OS and OP-Match who construct outlier detection would set all predicted outlier weights as 0 and the rest inlier as 1, thus in the top-right OP-Match weights figure, all weights are predicted as 0 or 1. Although the number of samples with a weight of 1 in $\set{U}$ is significantly more than that in $\set{W}$ and $\set{S}$, almost all weak and strong OOD samples and substantial amount of in-distribution unlabeled samples are excluded, resulting in a fewer training samples. 
Prediction by LTWA is shown in the bottom-left plot that it cannot distinguish $\set{U}$, $\set{W}$ and $\set{S}$ well. In our method, as shown in the bottom-right figure, weights are concentrated at either near 0 or 1. There are more weights near 1 in $\set{U}$, while more weights are close to 0 in $\set{W}$ and $\set{S}$. Therefore, the average of $\set{U}$ is significantly higher than $\set{W}$ and $\set{S}$ .
}

\begin{figure}[!t]
\centering   
\includegraphics[width=0.48\textwidth]{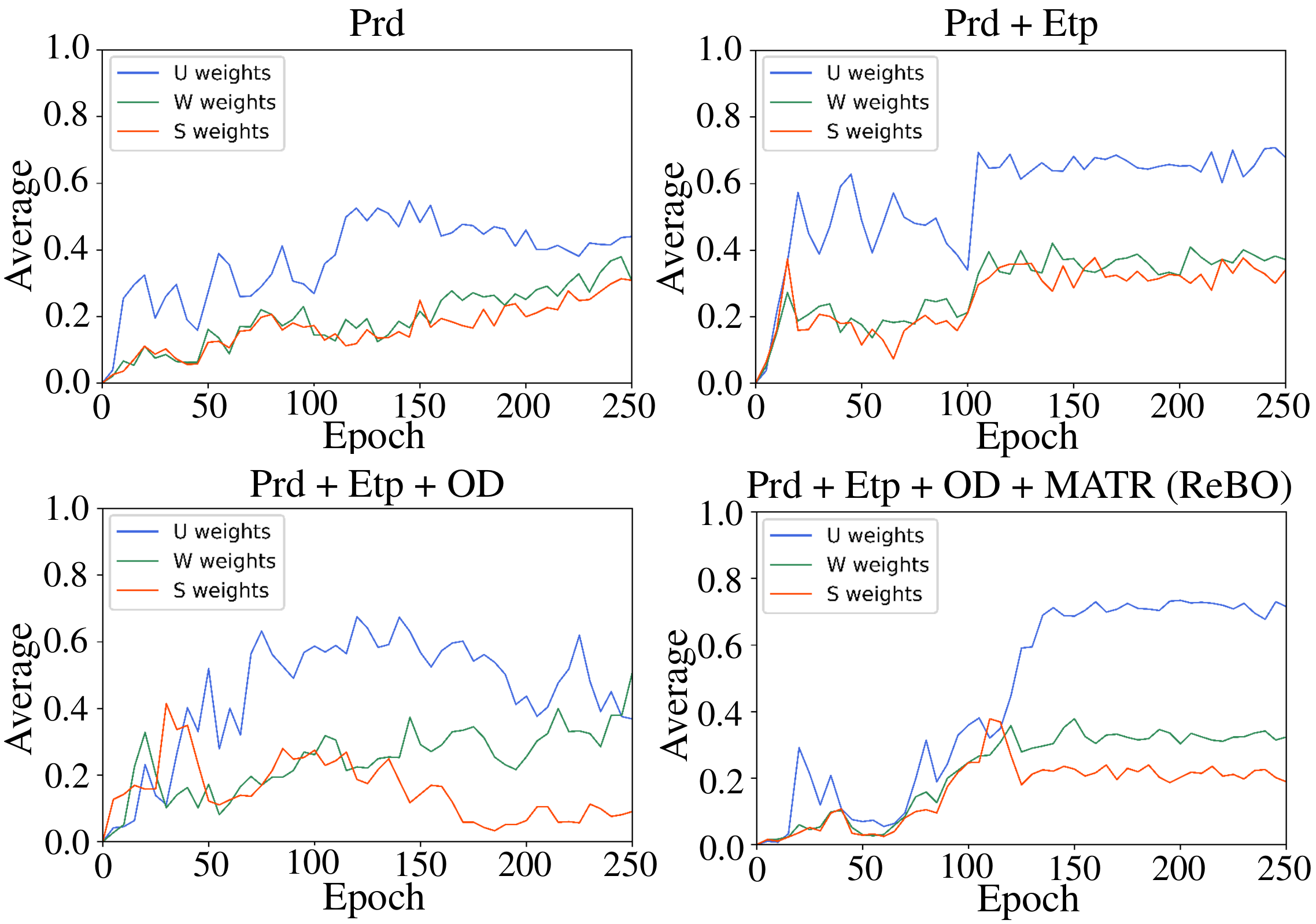}
\vspace{-0.5cm}
\caption{Predicted weights under different modules during training. Prd, Etp, OD, MATR refer to weight predictor, entropy loss, outlier detection, Moving Averaging Tikhonov regularization respectively.}
\label{Weights Diagram}
\vspace{-0.4cm}
\end{figure}

\section{Conclusion}
In this work, we are motivated by the negative impact of out-of-distribution unlabeled data and propose an open-set semi-supervised learning problem for 3D point cloud understanding. In light of the drawbacks of existing open-set SSL approaches, we propose a bi-level optimization based approach. The meta-objective is defined on a held-out validation set to avoid trivial solution. To further regularize the noisy meta-gradients, we propose three regularization terms, namely outlier detection, entropy regularization and moving averaging weights. Extensive experiments on 3D point cloud classification and segmentation validate the efficacy of proposed methods.



{\small
\bibliographystyle{ieee_fullname}
\bibliography{mian_arxiv}
}

\clearpage
\textbf{\Large Appendix}

\section{Continual Learning on Unseen Data}\label{sect:Unseen Generalization}
  {In a realistic open-set SSL environment the unlabeled data could continuously expand when more data is collected. Given the expensive computation cost of bi-level optimization, training from scratch when more unlabeled data becomes available is a very inefficient practice. Therefore, we consider conitnual learning on unseen data task in this section to simulate the ever expanding unlabeled dataset.
  Formally, we consider ShapeNet as an additional unseen dataset in our ModelNet40 classification experiments. When the backbone and weight predictor have been trained on ModelNet40($\set{L}$ + $\set{U}$) and S3DIS($\set{W}$ + $\set{S}$), we could estimate ShapeNet samples' weights directly without additional training. The weights on ShapeNet are fixed for continual training(50 epochs) of backbone. In addition, we could fine-tune our backbone and predictor for 50 epochs with additional unseen ShapeNet data. We explore these two options and present the results in Tab.~\ref{Generalization to Unseen Data}. It can be found that the model does benefit from additional unseen data with 1.9$\%$ @ 100 labeled and 0.6$\%$ @ 400 labeled improvements even without unlabeled data weighting. But further improvement is observed if we estimate and fix the unlabeled data weights (Est $\&$ Fix) , suggesting the benefits of selectively using unlabeled data. Finally, the best results is obtained by doing further fine-tuning (Fine-tune) with additional unseen data. This cheap fine-tuning approach turns out to be more effective.}

  \begin{table}[!htb]
    \caption{Transferring to Unseen Data. Uniform Unseen Weights means all weights in unseen data are set to 1. Est \& Fix and Fine-tune are two ways of our continual training.}
    \label{Generalization to Unseen Data}
    \centering
    \scalebox{1.0}{
    \begin{tabular}{c c c c}
      \toprule
      Methods & 100 labeled & 400 labeled \\
      \hline
      w/o Unseen Data & $63.4$ & $80.9$ \\
      Uniform Unseen Weights & $65.3$ & $81.5$ \\
        \hline
      Est \& Fix & $66.3$ & $82.1$ \\
      Fine-tune & \bm{$67.1$} & \bm{$82.4$} \\
      \bottomrule
    \end{tabular}
    }
  \end{table}

  \section{Additional Backbone}
  {
  To prove that our weight predictor is also applicable to the latest task networks, we swap our backbone with CurveNet \cite{xiang2021walk}, a point cloud shape analysis network proposed to obtain local information by curve walking. We first repeat ModelNet40 classification experiments and present results in Tab.~\ref{CurveNet Classification}. It shows that our weight predictor method still outperforms all other methods with state-of-the-art backbone network, reaching 69.1$\%$ / 81.4$\%$ accuracy under 100 / 400 labeled samples setting. We further explore the universality of weights across different backbones. We fix and transfer the weights estimated by DGCNN to CurveNet SSL training. The result (weight transfer) proves that CurveNet could work with the transferred weights and it is only slightly behind training from scratch. This indicates estimating unlabeled data weights with a light-weight model is a feasible way to save computation cost.
  }

\begin{table}[!t]
    \caption{Evaluation of our method conducted with CurveNet backbone on ModelNet40 classification.}
    \label{CurveNet Classification}
    \centering
    \scalebox{0.85}{
    \begin{tabular}{c c c c}
      \toprule
      Data & Methods & 100 labeled & 400 labeled\\
      \hline
      \multirow{7}{*} {$\{\set{L},\set{U},\set{W},\set{S}\}$} & Baseline & $67.2$ & $79.9$ \\
      & DS3L \cite{guo2020safe} & $67.8$ & $79.6$ \\
      & Multi-OS \cite{yu2020multi} & $67.4$ & $80.1$ \\
      & LTWA \cite{ren2020not} & $68.0$ & $79.9$ \\
      & OP-Match \cite{saito2021openmatch} & $67.8$ & $79.7$ \\
      \cline{2-4}
      & ReBO(weight transfer) & $68.3$ & $80.7$ \\
      & ReBO(train from scratch) & \bm{$69.1$} & \bm{$81.4$} \\
      \bottomrule
    \end{tabular}
    }
  \end{table}

\section{Temporal Evolution of Predicted Weights}
  {We analyze and visualize predicted weights on 100 labeled ModelNet40 ($\set{L},\set{U}$) classification experiment with additional S3DIS ($\set{W},\set{S}$). In Fig.~\ref{Weights Diagram}, we show the weights of $\set{U}$, $\set{W}$ and $\set{S}$ under our different modules over training epochs. The x-axis is the number of training epoch. It can be seen from the top-left plot that when no regularization constraint is applied, although predictor assigns higher weights to $\set{U}$, it fails to converge. It is because in the later stage of training, the converged classification network will receive weaker gradients from unsupervised loss, thus producing unstable meta-gradient. The top-right plot shows that entropy loss, which forces predicted weights to 0 or 1, could stabilize predictor in the later stage of training. In the bottom-left plot, applying outlier detection could effectively distinguish $\set{W}$ and $\set{S}$, as it is conducive to focusing predictor on object feature. However, it results in increasing $\set{W}$ weights in the end. The plot at the bottom-right shows the effect of the final model, which significantly smooths and stabilizes the predicted weights. In the end, predictor could reasonably distinguish $\set{U}$, $\set{W}$ and $\set{S}$, assigning higher weights to $\set{U}$ and lower weights to $\set{S}$.
  }

  \begin{figure}[!t]
  \centering   
  \includegraphics[width=0.48\textwidth]{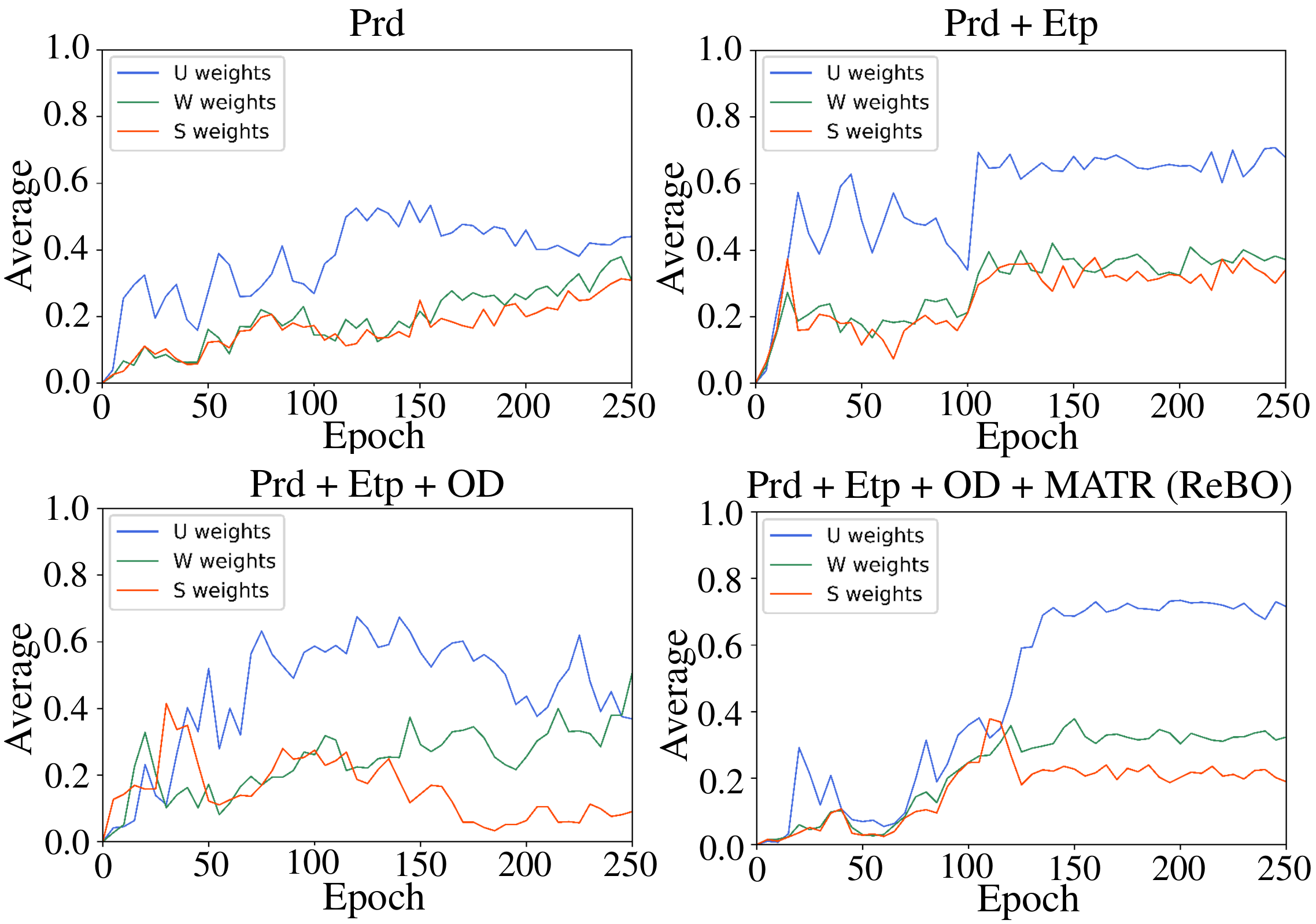}
  \caption{Predicted weights under different modules during training. Prd, Etp, OD, MATR refer to weight predictor, entropy loss, outlier detection, Moving Averaging Tikhonov regularization respectively.}
  \label{Weights Diagram}
  \end{figure}

\section{Finite Difference Approximation}
\label{sect:Finite Difference Approximation}
In this section, we provide the derivations for approximating the meta-gradient with finite difference approximation in Eq.~(4) of the manuscript. We first write the approximated meta-gradient with one-step gradient descent as,
\begin{equation}
    \nabla_{\Phi}\mathcal{L}_{val}=-\alpha\nabla_{\Phi,\Theta}^2\mathcal{L}_{tr}(\Theta,\Phi)\nabla_{\Theta^*}\mathcal{L}_{val}(\Theta^*,\Phi)
\end{equation}

To efficiently compute the hessian-vector product, we adopt the finite difference approximation. In the following, we prove that the finite difference approximation can be achieved with the help of a linear approximation of training loss gradient w.r.t. hyperparameters.

W.l.o.g, let $g(\Theta,\Phi)=\nabla_{\Phi}\mathcal{L}_{tr}(\Theta,\Phi)$ and $\vect{v}=\nabla_{\Theta^*}\mathcal{L}_{val}(\Theta^*,\Phi)$.

Through Taylor series at $\Theta$ we have the following equality,
\begin{equation}
    g(\Theta+\epsilon\vect{v})=g(\Theta) + g^\prime(\Theta)\epsilon\vect{v} + o(\epsilon^2\vect{v}^2)
\end{equation}

When $\epsilon\rightarrow 0$ we have,
\begin{equation}
    g(\Theta+\epsilon\vect{v})\approx g(\Theta) + g^\prime(\Theta)\epsilon\vect{v}
\end{equation}

From the equality above, we have,
\begin{equation}\label{eq:g_prime}
    g^\prime(\Theta)\vect{v}=\frac{g(\Theta+\epsilon\vect{v})-g(\Theta)}{\epsilon}
\end{equation}

Substituting $g(\Theta)$ and $\vect{v}$ in Eq.~(\ref{eq:g_prime}), we have,
\begin{equation}\label{eq:fwd_fda}
\begin{split}
    &\nabla_{\Phi,\Theta}^2\mathcal{L}_{tr}(\Theta,\Phi)\nabla_{\Theta^*}\mathcal{L}_{val}(\Theta^*,\Phi)=\\
    &\frac{\nabla_{\Phi}\mathcal{L}_{tr}(\Theta+\epsilon\nabla_{\Theta^*}\mathcal{L}_{val}(\Theta^*,\Phi),\Phi)-\nabla_{\Phi}\mathcal{L}_{tr}(\Theta,\Phi)}{\epsilon}
\end{split}
\end{equation}

Eq.~(\ref{eq:fwd_fda}) can be interpreted as a forward finite difference approximation with a step size of $\epsilon$. A more accurate approximation would be a central difference and then we arrive at the approximation result in Eq~(4) of the manuscript as,

\begin{equation}
\resizebox{0.99\linewidth}{!}{
	$
    \begin{split}
    &\nabla_{\Phi,\Theta}^2\mathcal{L}_{tr}(\Theta,\Phi)\nabla_{\Theta^*}\mathcal{L}_{val}(\Theta^*,\Phi)=\\
    &\frac{\nabla_{\Phi}\mathcal{L}_{tr}(\Theta+\epsilon\nabla_{\Theta^*}\mathcal{L}_{val}(\Theta^*,\Phi),\Phi)-\nabla_{\Phi}\mathcal{L}_{tr}(\Theta-\epsilon\nabla_{\Theta^*}\mathcal{L}_{val}(\Theta^*,\Phi),\Phi)}{2\epsilon}
\end{split}
$
}
\end{equation}

\section{Piecewise Increasing Weight for Entropy Loss}
\label{sect:Hyper-Parameter of Entropy Loss}
Entropy Loss works by encouraging weight predictor to produce close to binary weights for unlabeled samples.
Encouraging low entropy on unlabeled data weights too early may harm the learning and potentially results in trivial solutions.
Thus, in ReBO, we adopt a warm-up weight curve for entropy loss to gradually increase the strength of entropy loss. In specific, weight $\xi$ is designed as a piecewise increasing function to control the strength of entropy loss during training.
\begin{equation}
\begin{cases}
\xi = 0, & \text{ if } e_t<50 \\ 
\xi =\delta * \frac{e_t - 50}{100}, & \text{ if } 50<=e_t<=100 \\ 
\xi = \delta, & \text{ if } e_t>100
\end{cases}
\end{equation}
where $e_t$ is the number of training epochs and $\delta$ is a hyper-parameter which will be discussed in the next section.


\begin{figure*}[!t]
\centering   
\includegraphics[width=0.8\textwidth]{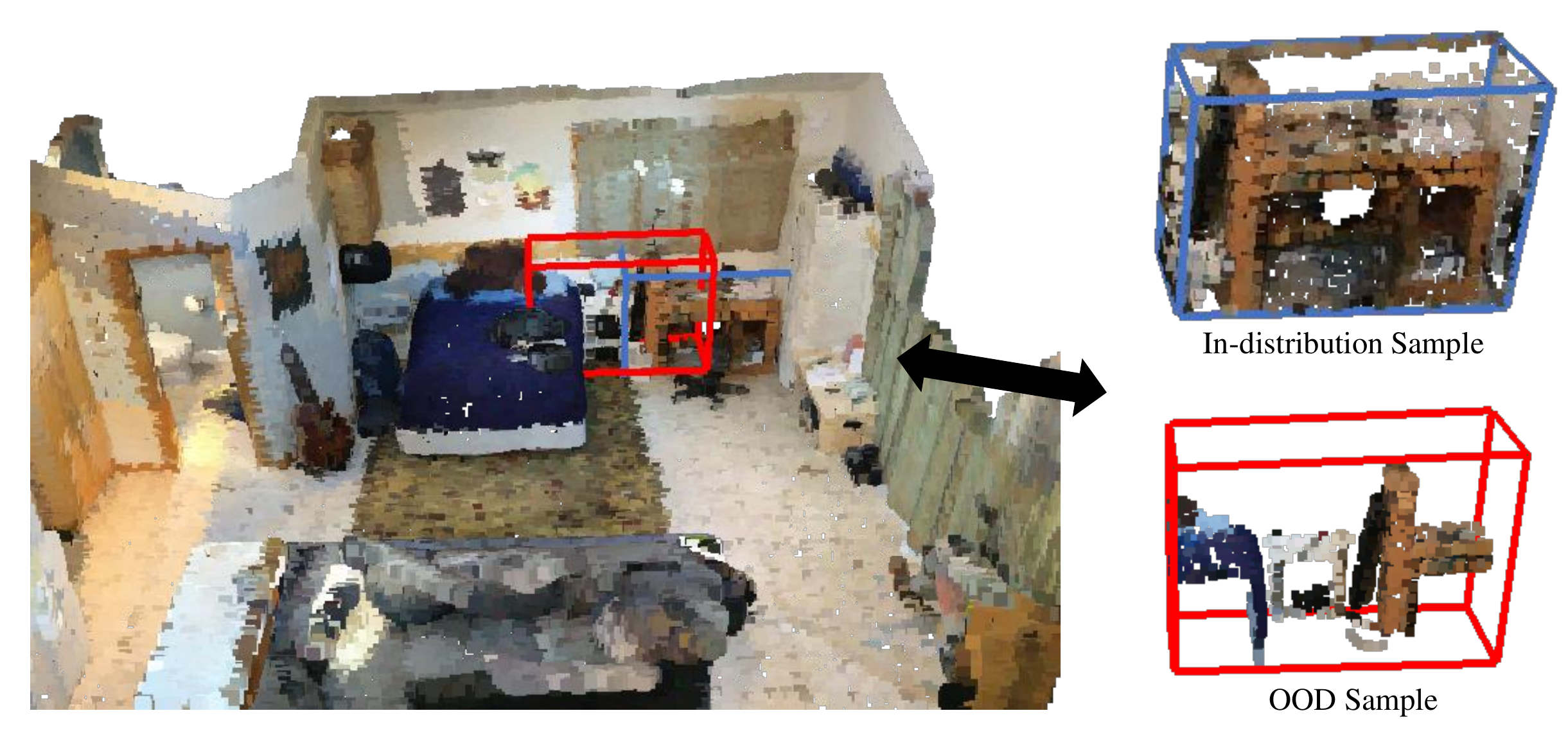}
\caption{Point cloud in the blue box belongs to in-distribution. We transform the blue box in position and size to create the red box, and then crop out the OOD sample from it. 
}
\label{Real Scan OOD}
\end{figure*}

\section{Additional Evaluation of Hyper-Parameters}
\label{sect:Evaluation of Hyper-Parameters}
{In this section, we present additional evaluation of the selection of $\delta$ and other hyper-parameters in Eq.~(7) and Eq.~(9) of the manuscript. We conduct the experiments on 100 labeled samples $\set{L}$ and the rest $\set{U}$ from ModelNet40 classification with $\set{W}$ and $\set{S}$ from S3DIS. In Tab.~\ref{Hyper-Parameters Evaluation}, under default setting ($\beta$=0.5, $\gamma$=0.1, $\eta$=0.01, $\delta$=0.01), we separately evaluate each individual hyperparameter and present the comparisons in Tab.~\ref{Hyper-Parameters Evaluation}.}

\begin{table}[!htb]
	\caption{Evaluation of Hyper-Parameters Selection. Hyper-parameters $\beta$, $\gamma$, $\eta$, $\delta$ are set as 0.5, 0.1, 0.01, 0.01 by default and we change them one by one in the experiments.}
	\label{Hyper-Parameters Evaluation}
	\centering
	\scalebox{0.8}{
	\begin{tabular}{c c | c c | c c | c c}
		\toprule
		$\beta$ & Acc(\%) & $\gamma$ & Acc(\%) & $\eta$ & Acc(\%) & $\delta$ & Acc(\%) \\	
		\hline
		0.1 & $62.8$ & 0.01 & $62.7$ & 0.001 & $61.9$ & 0.001 & $62.0$ \\
		0.5 & \bm{$63.4$} & 0.1 & \bm{$63.4$} & 0.01 & \bm{$63.4$} & 0.01 & \bm{$63.4$} \\
		0.9 & $61.6$ & 1.0 & $61.3$ & 0.1 & $62.3$ & 0.1 & $60.8$  \\
		\bottomrule
	\end{tabular}
	}
\end{table}

\section{Generating OOD Samples for Real Scan Data}
\label{sect:Generating Noisy OOD Samples}
To generate more realistic OOD samples for real scan data, we consider perturbing object bounding boxes in ScanNet and SceneNN, and then crop out point clouds from perturbed boxes as OOD samples. In more details, for a bounding box with center $\{x,y,z\}$ and size $\{l_x,l_y,l_z\}$, we transform it to a new box $\{x',y',z',{l_x}',{l_y}',{l_z}'\}$ by
\begin{equation}
\begin{split}
    \begin{cases}
    t' = t + \alpha l_t \\
    {l_t}' = (1 + \beta) l_t
    \end{cases} 
    \forall{\scal{t}} \in \{\scal{x}, \scal{y}, \scal{z}\} 
\end{split}
\end{equation}
where $\alpha$ is uniformly sampled from $\left [ -1.0, -0.5 \right ] \cup \left [ 0.5, 1.0  \right ]$, ensuring a limited overlap between OOD samples and In-distribution samples, and $\beta$ follows zero-mean Gaussian distribution with 0.2 standard deviation. 
In Fig.~\ref{Real Scan OOD}, we display an example for real scan OOD point cloud, where we disturb the box of an in-distribution sample(a desk in blue box) and obtain the OOD sample from the new box(an OOD sample in red box). 

\section{Visualization of Weighted Samples}
\label{sect:Visualization of Weighted Samples}
\begin{figure*}[!htb]
\centering   
\includegraphics[width=1.0\textwidth]{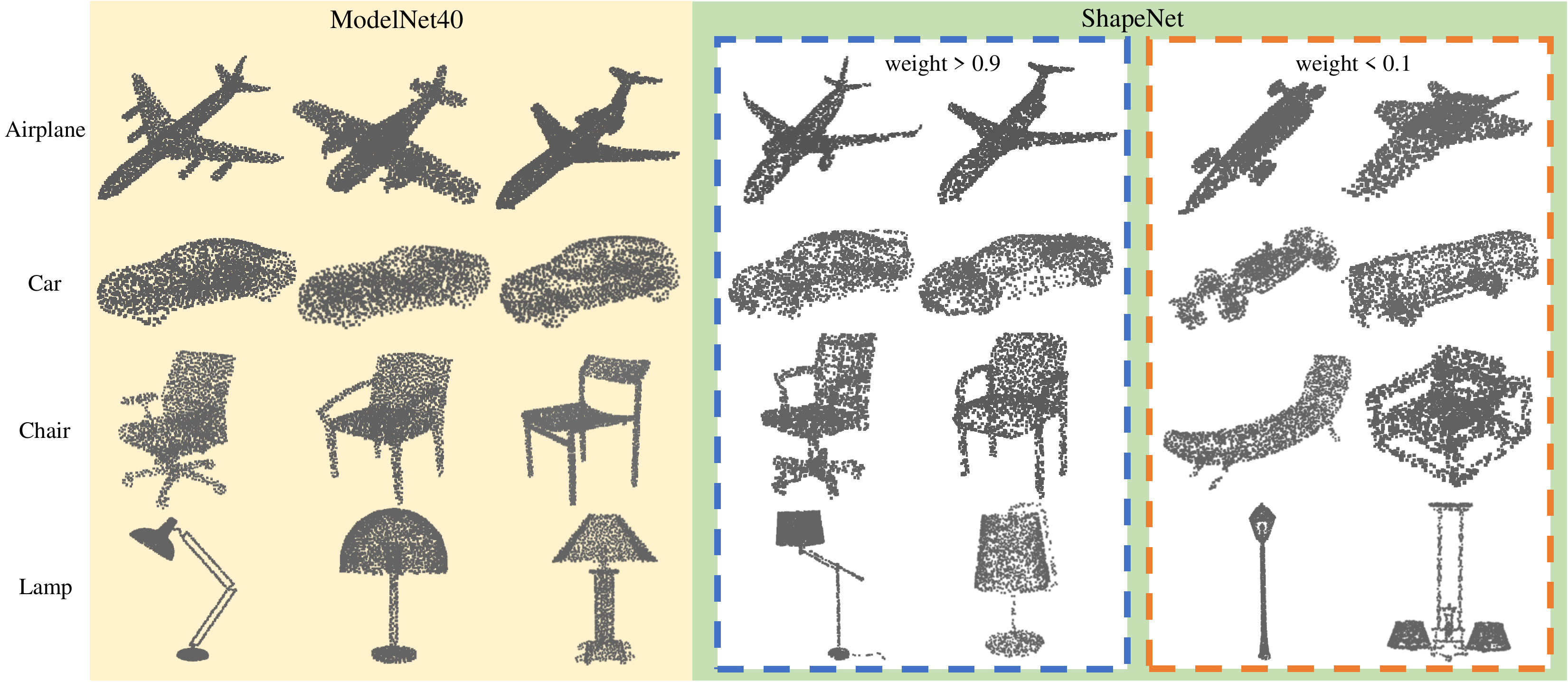}
\caption{Labeled samples from ModelNet40 are on the left and ShapeNet samples with the same semantic labels are on the right. The predicted weights of samples indicate the whether these unlabeled data are conducive to target task or not.
}
\label{Weighted Samples}
\end{figure*}

To explore which unlabeled samples are prioritized in our method, we visualize some samples in Continual Learning on Unseen Data experiments, where backbone and weight predictor have been trained on ModelNet40~(400 labeled $\set{L}$ + the rest $\set{U}$) and S3DIS~($\set{W}$ + $\set{S}$), and ShapeNet is considered as an additional unseen dataset continual training. Although there are some overlapping categories in both ModelNet40 and ShapeNet, the distributions of data samples across two datasets in the same semantic category do not match exactly. 
In Fig.~\ref{Weighted Samples}, we visualize selected samples of four semantic categories from both datasets. On the left, there are labeled point cloud samples from ModelNet40. On the right, we observe samples which are visually similar, i.e. similar global topology and local geometry, to ModelNet40 data are given higher weights (weight$>$0.9). While samples that are not visually similar are given lower weights despite labeled with the same semantic category (weight$<$0.1). These observations suggest ReBO has the ability to selectively exploit unlabeled data to improve the performance on target dataset.


\section{Qualitative Results for Part Segmentation}
\label{sect:Qualitative Results for Part Segmentation}
\begin{figure*}[!htb]
\centering   
\includegraphics[width=1.0\textwidth]{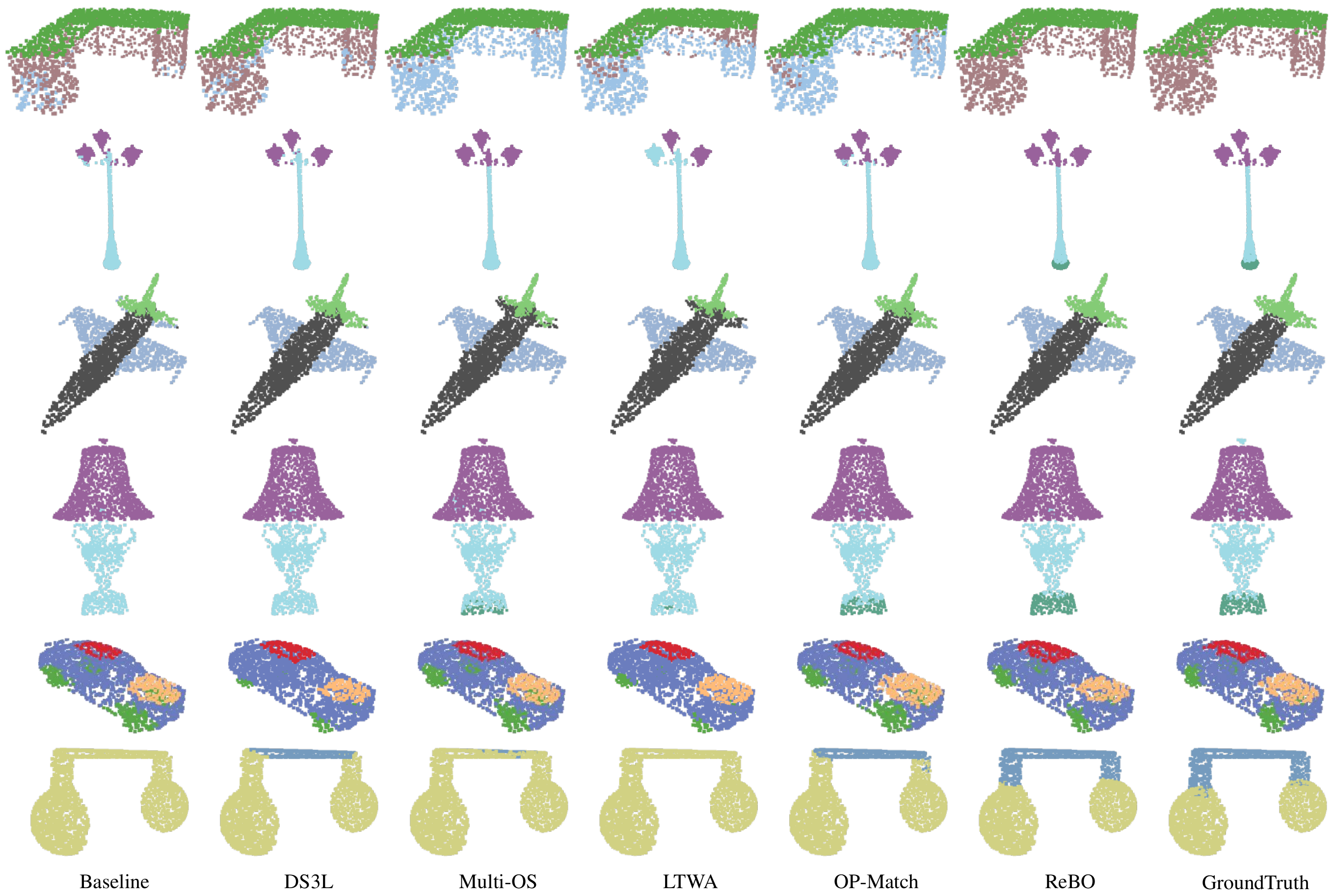}
\caption{Part segmentation results among different methods are presented. On the far right is the semantic ground truth.}
\label{partseg results}
\end{figure*}

In this section, we present the qualitative results of ShapeNet part-segmentation experiments carried on 400 labeled samples $\set{L}$, the rest unlabeled samples $\set{U}$ and additional OOD S3DIS samples $\set{W}$ and $\set{S}$. As shown in Fig.~\ref{partseg results}, compared with other methods, our ReBO achieves more accurate segmentation, which is closer to semantic ground truth, especially on some small parts like the base of the lamp and the tires of the car.

\section{Performance Gap at Low-Label Regime}
\begin{figure*}[!htb]
\centering   
\includegraphics[width=1\linewidth]{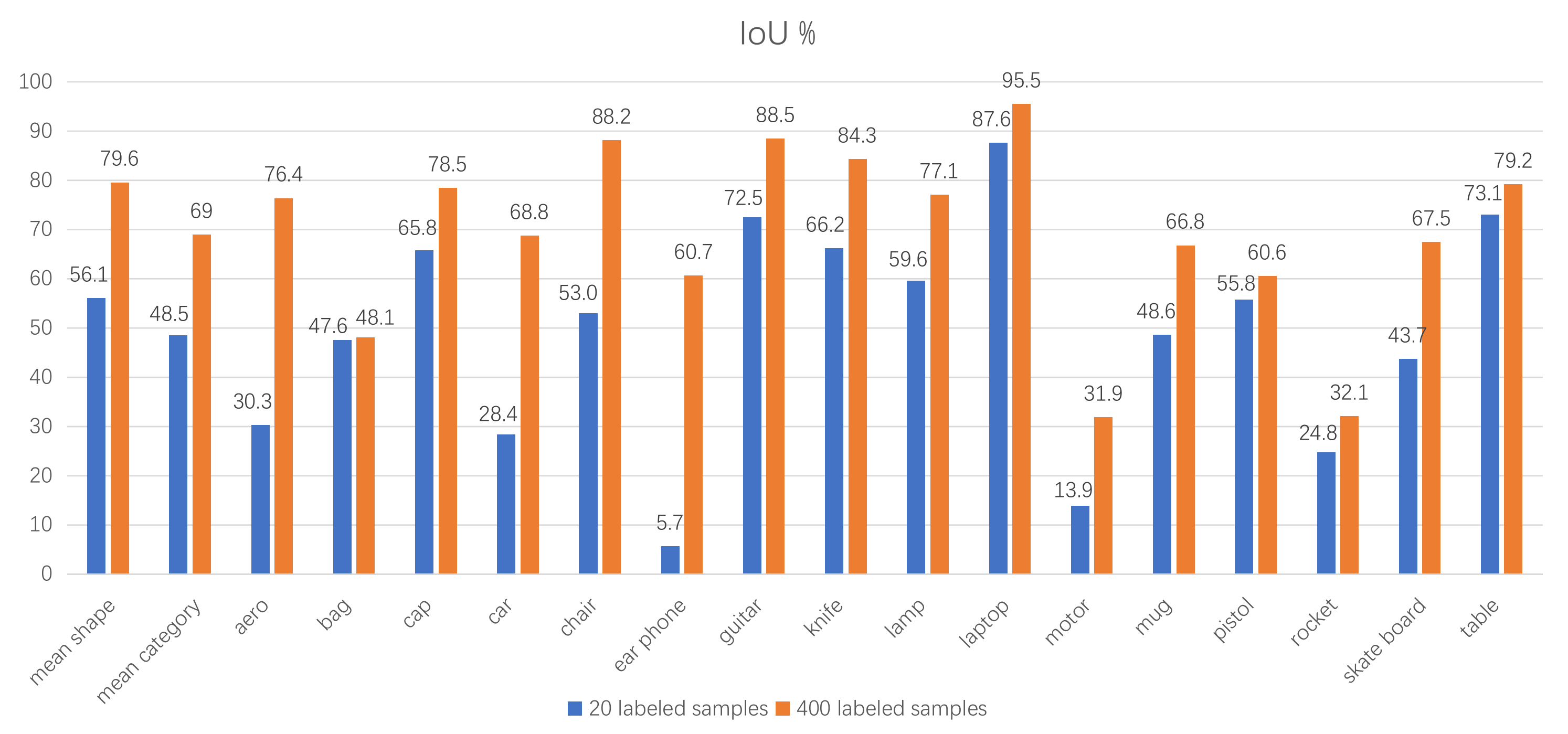}
\caption{ShapeNet part segmentation result under 20 labeled training samples and 400 labeled training samples.}
\label{iou histogram}
\end{figure*}

We observe a significant performance gap between 20 labeled data and 400 labeled data on ShapeNet segmentation dataset and attribute the performance gap to the following reasons. Firstly, mIoU at 0.54 is a relatively poor result compared against mIoU at 0.90 when all training samples are labeled. This suggests segmentation at extremely low labeled data is still a very challenging task. Secondly, some categories in ShapeNet dataset has smaller intra-class variation, e.g. most samples in the plane category are rather similar. Therefore, 1-2 examples per-class enables training a reasonably well network. But againt, the performance at 20 labeled samples is still significantly worse than the fully supervised results. In Fig.~\ref{iou histogram}, we show out each ShapeNet Part category iou by ReBO trained on 20 labeled samples and 400 labeled samples respectively. It can be seen that the performance of some categories, such as ear phone, is really poor (only $5.7\%$ iou) with only 20 labeled samples, but improved significantly (up to $60.7\%$ iou) when the number of labeled samples increased to 400. On the other hand, there was no significant difference between 20labeled and 400 labeled. On the other hand, for some categories, such as bag, there is no significant gap between the results with 20 labeled and 400 labeled ($47.6\%$ and $48.1\%$ iou).

\section{Loss Progression}

\begin{figure}[!htb]
\centering   
\includegraphics[width=1\linewidth]{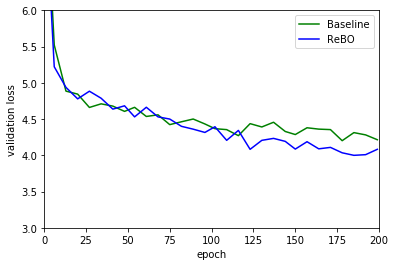}
\caption{Loss progression curves during ModelNet40 classification training.}
\label{loss curve}
\end{figure}

In Fig.~\ref{loss curve}, we present the validation loss curve on ModelNet40 classification with 100 labeled samples for the baseline method and our proposed ReBO. It demonstrates that with the help of weight predictor, the model could converge to a better solution, i.e. lower validation loss.

\end{document}